# Risk Analysis in Customer Relationship Management via Quantile Region Convolutional Neural Network-Long Short-Term Memory and Cross-Attention Mechanism

Yaowen Huang[1,2], Jun Der Leu[2], Baoli Lu[3,4], Yan Zhou[5]

[1]Longyan University, Fujian, Longyan, 364012, China

[2]Department of Business Administration, National Central University, Taoyuan City, Taiwan, 32001, China

[3]University of Portsmouth, Portsmouth, PO1 2UP, United Kingdom

[4]Institute of Semiconductors, Chinese Academy of Sciences, Beijing, 100083, China

[5]Northeastern University, San Jose, California, 95131, United States

## Abstract

Risk analysis is an important business decision support task in customer relationship management (CRM), involving the identification of potential risks or challenges that may affect customer satisfaction, retention rates, and overall business performance. To enhance risk analysis in CRM, this paper combines the advantages of quantile region convolutional neural network-long short-term memory (QRCNN-LSTM) and cross-attention mechanisms for modeling. The QRCNN-LSTM model combines sequence modeling with deep learning architectures commonly used in natural language processing tasks, enabling the capture of both local and global dependencies in sequence data. The cross-attention mechanism enhances interactions between different input data parts, allowing the model to focus on specific areas or features relevant to CRM risk analysis. By applying QRCNN-LSTM and cross-attention mechanisms to CRM risk analysis, empirical evidence demonstrates that this approach can effectively identify potential risks and provide data-driven support for business decisions.

business decision

## Introduction

In today's competitive business environment, customer relationship management (CRM) is one of the key factors for success (Haiyun et al., 2021). To provide personalized services, increase customer satisfaction, and improve sales performance, businesses need to deeply understand and analyze customer behavior and timely identify potential risks (Li et al., 2020b). Traditional risk analysis methods often rely on experience and intuition, but the development of deep learning and machine learning technologies offers new opportunities and challenges for risk analysis in CRM (Libai et al., 2020).

Some commonly used deep learning or machine learning models are:

- Logistic regression model (Guerola-Navarro et al., 2021): Logistic regression is a widely used classification algorithm that can predict different risk categories. However, logistic regression models often fail to capture complex nonlinear relationships.
- Decision tree model (Chen et al., 2021): Decision tree models can generate easy-to-understand rules and have good interpretability. But they tend to overfit when dealing with data that has complex structures and high-dimensional features.
- Random forest model (Rao et al., 2020): Random forests are an ensemble learning method that improves prediction performance by combining multiple decision tree models. However, the computational complexity of random forests is high, making them less suitable for large datasets.
- Convolutional neural network (CNN) model (De Caigny et al., 2020): CNNs excel in image recognition and computer vision. However, in CRM risk analysis, CNN models may not adequately consider the temporal information of sequential data.
- Long short-term memory (LSTM) model (Goel & Bajpai, 2020): LSTMs are a type of recurrent neural network suitable for sequence modeling, effectively handling long-term dependencies. However, LSTM models may fail to capture local feature information in sequence data.

The three most relevant directions related to this theme include:

- Advanced risk analysis models: Researchers can further explore and develop more advanced deep learning or machine learning models to address more complex risk analysis tasks in CRM (Ponzo et al., 2021). For example, considering models with attention mechanisms, graph neural networks, or generative adversarial networks could be beneficial to capture finer-grained risk features and handle multimodal data risk analysis (Zhang et al., 2022).
- Real-time and personalized risk prediction: Researchers can focus on how to integrate real-time data sources (Chakraborty & Chosh, 2020) (such as social media data, real-time interaction data, etc.) with customer data in CRM systems to achieve more timely and personalized risk predictions (Li et al., 2020a). This may involve streaming data processing, incremental learning, and online learning technologies to monitor and predict risks in real-time and provide personalized risk analysis results and recommendations based on individual customer characteristics (Fazakis et al., 2021).
- Integrating domain knowledge and interpretative analysis: Risk analysis in CRM often requires integration of domain experts' knowledge and experience for interpretation and decision support (Han et al., 2020). Therefore, researchers can explore how to merge deep learning and machine learning models with domain knowledge to improve model interpretability and explainability (Kumar & Goel, 2022). Additionally, developing visualization and interactive tools to help users understand and interpret the model's predictions, and to effectively communicate and collaborate with domain experts, is another important research direction (Nelson et al., 2022).

Considering the limitations of these models, this study aims to propose a method that combines a quantile region convolutional neural network (QRCNN)-LSTM with a cross-attention mechanism to enhance risk analysis in CRM, considering the limitations of these models. The QRCNN-LSTM model is capable of capturing both local and global dependencies in sequence data, thereby improving the accuracy of risk probability prediction. By incorporating a cross-attention mechanism, the model can effectively focus on essential features relevant to risk analysis, resulting in enhanced precision and robustness.

The proposed method follows a sequential process. Firstly, the QRCNN-LSTM model extracts local features using four local convolutional kernels. Subsequently, LSTM layers are employed to handle long-term dependencies in the sequence data. The cross-attention mechanisms are then employed to assign weights to different local features, facilitating the capture of global information. Finally, the weighted features are fed into the output layer for risk probability prediction.

Experimental results have demonstrated significant performance improvements in CRM risk analysis tasks with the proposed method. When compared to traditional models, the combination of QRCNN-LSTM with a cross-attention mechanism exhibits superior accuracy in identifying potential risks, thereby providing more reliable data support for business decision-making. The proposed method, which combines QRCNN-LSTM with a cross-attention mechanism, holds promise for CRM risk analysis. By leveraging local feature extraction, global dependency modeling, and a weighted attention mechanism, this method effectively captures risk features and enhances prediction accuracy. This is of great significance for businesses seeking to optimize customer relationships and improve overall business performance.

In conclusion, this research aims to enhance risk analysis in CRM by proposing a method based on QRCNN-LSTM and a cross-attention mechanism. By effectively capturing both local and global dependencies in sequence data and focusing on important features relevant to risk analysis, this method improves the accuracy and reliability of risk predictions, providing valuable support for business decision-making. A method based on QRCNN-LSTM and a cross-attention mechanism has been proposed to improve risk analysis in CRM. By integrating the sequence modeling capabilities of the QRCNN-LSTM model with the advantages of natural language processing, and the weighted focus ability of the cross-attention mechanism, this method can more accurately identify potential risks, providing data-driven decision support.

- Experimental results have shown that the proposed method achieves significant performance improvements in CRM risk analysis tasks. Compared to traditional models such as logistic regression, decision trees, random forests, and convolutional neural networks, the combination of QRCNN-LSTM and the cross-attention mechanism better captures local and global dependencies in sequence data, enhancing the accuracy and reliability of risk predictions.
- The outcomes of this study are significant for improving customer satisfaction, increasing retention rates, and optimizing business performance. By accurately identifying and assessing potential risks, businesses can take timely measures, provide personalized services, and strengthen customer relationships. This, in turn, enhances customer satisfaction and retention rates, ultimately boosting overall business performance.

## Methodology

### Overview of the Network

This method proposes a combination of the QRCNN-LSTM model and a cross-attention mechanism to enhance risk analysis tasks in CRM. The QRCNN-LSTM model leverages sequence modeling and natural language processing techniques to capture dependencies in sequential data. The cross-attention mechanism further improves the model's ability to identify risks by focusing on important features. In experiments, this method has demonstrated significant performance improvements, which are crucial for enhancing customer satisfaction and optimizing business performance. The proposed model's overall framework is depicted in Figure 1.

**Figure 1**

*The Overall Framework Diagram of the Proposed Model*

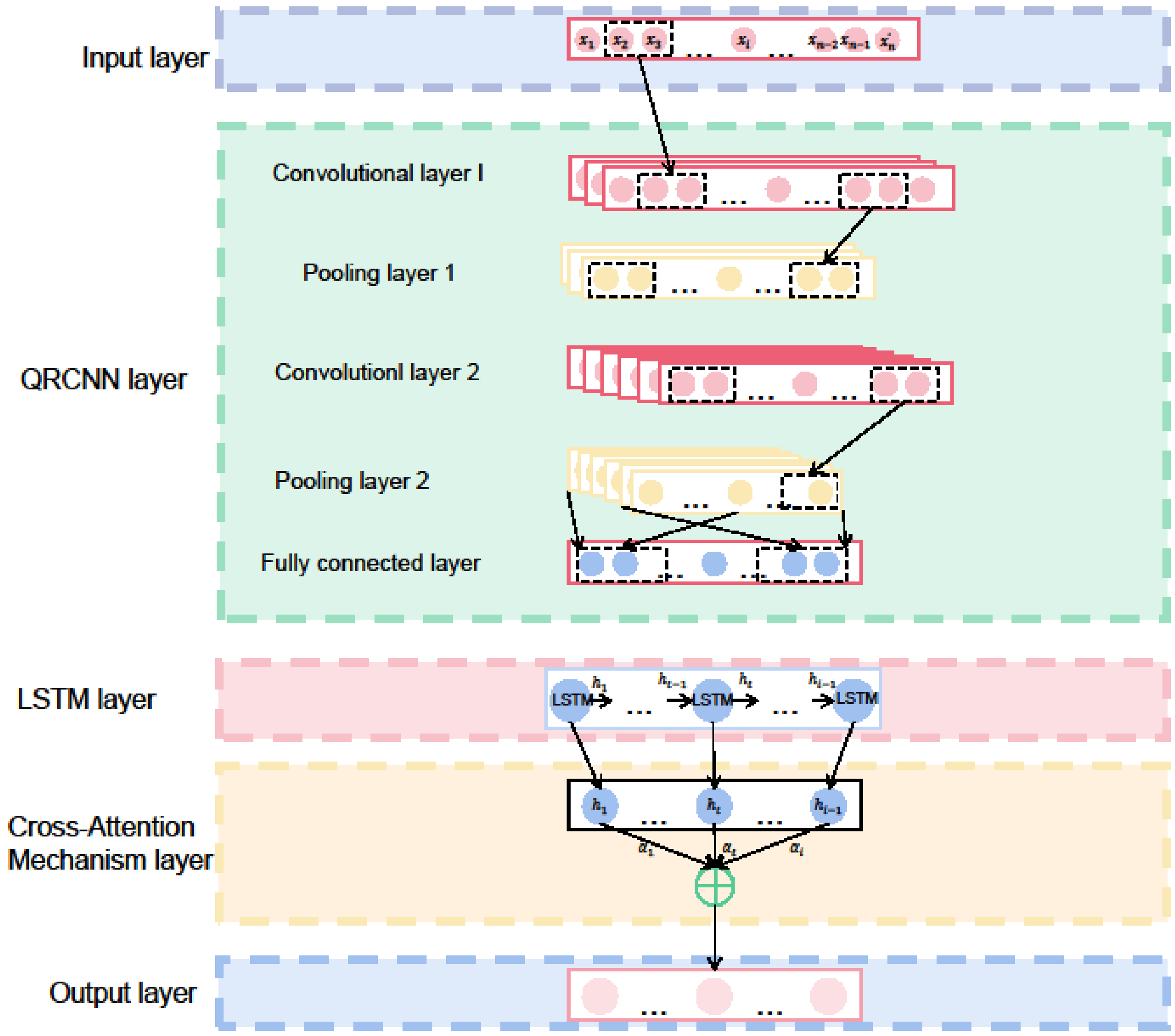

The core of the method is the QRCNN-LSTM model, which comprises two main components: the QRCNN and the LSTM network. The QRCNN model extracts local features from the input sequence using a CNN and combines them into a fixed-length representation using quick region pooling. The LSTM model then takes the output of the QRCNN as input and learns long-term dependencies within the sequence. By integrating QRCNN and LSTM, the model effectively captures both local and global dependencies in the sequential data. In the context of risk analysis, the model's input consists of various customer-related data such as customer behavior records, transaction history, social media information, etc. This data undergoes preprocessing and feature extraction steps to fit the input format required by the QRCNN-LSTM model. During the training phase, the model learns risk patterns in the data by minimizing the loss function and optimizing its parameters. Once trained, the model can be applied to new and unknown data for risk prediction and analysis.

Overall, this method offers a promising approach to enhance risk analysis in CRM by combining the QRCNN-LSTM model with a cross-attention mechanism. By effectively capturing dependencies and focusing on relevant features, the model provides valuable insights for managing risks, ultimately improving customer satisfaction and optimizing business performance.

The overall implementation process is as follows:

- Data Preprocessing: The first step involves cleaning and normalizing the raw data, removing any inconsistencies or outliers. The data is then processed to extract relevant features that will be used as inputs for the model. This includes transforming categorical variables into numerical representations and scaling numerical features if necessary.
- Building the QRCNN-LSTM Model: The QRCNN-LSTM model is constructed by defining its architecture, including the number of layers, the number of neurons in each layer, and the activation functions to be used. The QRCNN component extracts local features from the input sequence, while the LSTM component captures long-term dependencies.
- Model Training: The QRCNN-LSTM model is trained using a labeled historical dataset. During training, the model learns to minimize the loss function by adjusting its parameters through backpropagation and optimization algorithms like stochastic gradient descent. The goal is to optimize the model's ability to predict risk levels accurately.

- Model Evaluation: The trained model is evaluated using an independent test dataset that was not used during training. Performance metrics such as accuracy, recall, and F1-score are calculated to assess the model's effectiveness in predicting risks. This evaluation helps determine the model's reliability and generalization capability.
- Risk Prediction and Analysis: Once the model is trained and evaluated, it can be applied to new, unknown data for risk prediction and analysis. The model takes the input data and generates risk levels or probabilities as outputs. Businesses can use these predictions to identify potential risks and make informed decisions to mitigate them.

The construction and training phase of the QRCNN-LSTM model are critical in ensuring the method's performance. Proper parameter choices and model architecture are essential for accurate risk analysis. Additionally, data preprocessing and model evaluation play significant roles in ensuring the effectiveness and reliability of the method. By following this overall process, the method can provide accurate and reliable risk analysis results, offering valuable support for business decision-making.

**QRCNN Network**

The QRCNN model (Xu et al., 2023) is a deep learning model that combines the principles of CNNs and quantile regression. CNN is a deep learning model originally designed for image processing. Its fundamental principle involves performing convolution operations using filters (also known as kernels) on input data to capture local features within the data. This makes CNNs particularly suitable for processing data with spatial or time-series structures (Wang et al., 2022). Quantile regression is a regression analysis method that not only focuses on estimating the mean (or median) of the data but also considers various quantiles of the data distribution (e.g., 25th percentile, 75th percentile, etc.). Quantile regression is capable of estimating the conditional distribution at different quantiles, making it useful for addressing data uncertainty and risk. Figure 2 is a schematic diagram of the principle of QRCNN.

**Figure 2**

*The Schematic Diagram of the Principle of QRCNN*

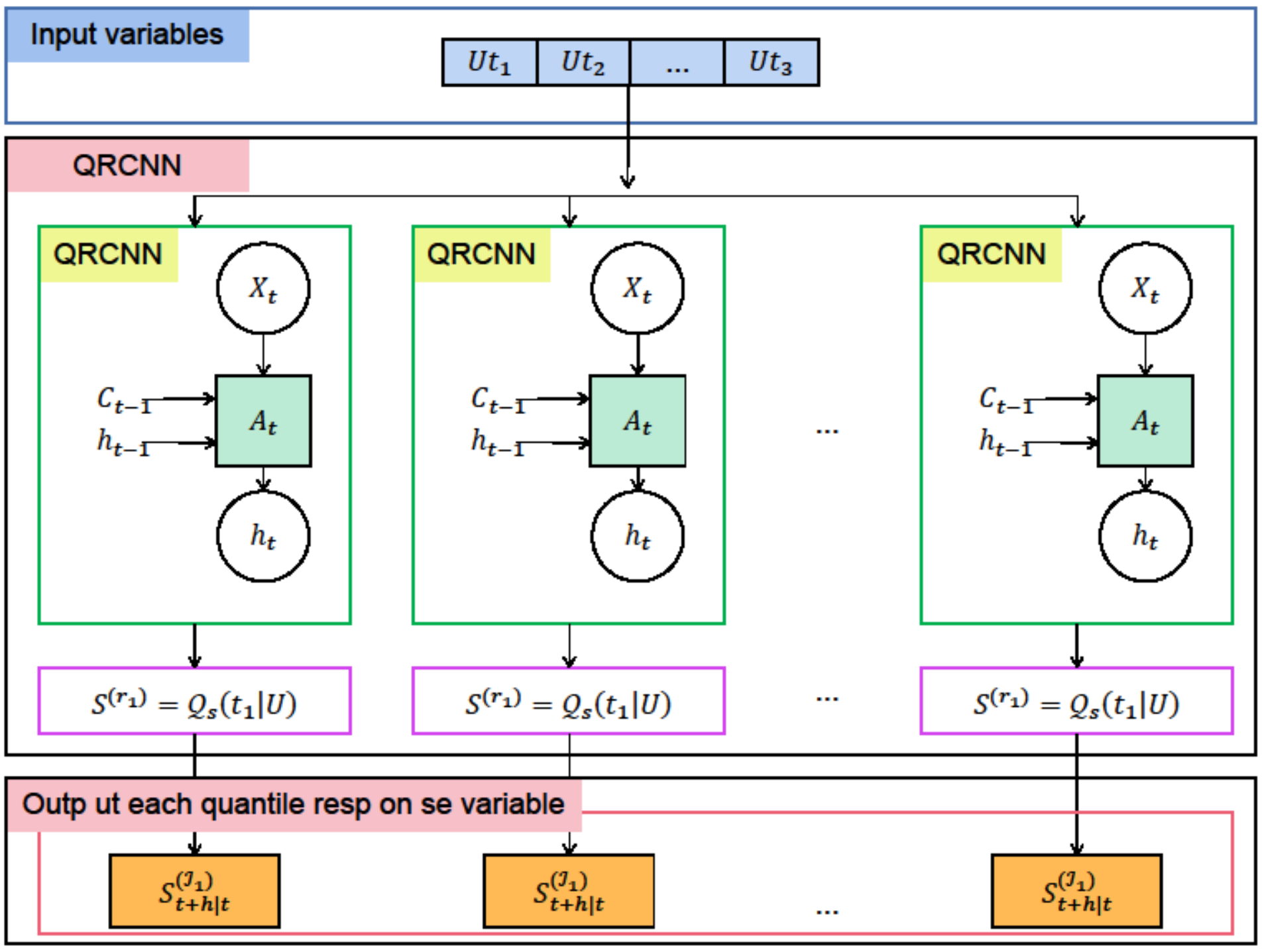

The QRCNN model, a variant of the CNN, is specifically designed to extract local features from input sequences using convolution operations. It introduces the concept of quick region pooling, which combines local features from different positions into a fixed-length representation.

The QRCNN model consists of multiple convolutional and pooling layers. Each convolutional layer performs local feature extraction by applying a sliding window over the input sequence, producing a set of feature maps. The quick region pooling layer then combines these feature maps to form a fixed-length vector representation, effectively capturing local features at various positions.

In this method, one-dimensional convolution operations are typically employed in the QRCNN model, with the input being sequence data such as text or time series. The size and stride of the convolutional kernels can be adjusted based on the specific task requirements. The quick region pooling layer commonly utilizes max pooling or average pooling operations to extract the maximum or average value from each feature map, resulting in a fixed-length feature vector.

During training, the QRCNN model learns the parameters of the convolutional and pooling layers through backpropagation and optimization algorithms like stochastic gradient descent. The objective is to maximize the model's performance in classification or regression tasks on the training data.

In the context of risk analysis tasks, the QRCNN model plays a crucial role in extracting important local features from customer-related input data, such as customer behavior patterns, transaction history, and social media information. These extracted features enable the identification of potential risk factors and provide valuable information for subsequent risk prediction and analysis.

The combination of the QRCNN and LSTM models is a key aspect of this method. While the QRCNN model focuses on extracting local features, the LSTM model captures long-term dependencies within the sequence data. This integration allows the model to consider both local and global information, enhancing the accuracy and reliability of risk analysis results.

Formula (1) is the formula of QRCNN:

$$\widehat{y_\tau} = f_\tau(x) = \beta_0^\tau + \sum_{j=1}^{J} \beta_j^\tau h_j(x) \tag{1}$$

Where:

$\widehat{y_\tau}$ means the estimated quantile is the predicted value at $\tau$, $x$ represents the input feature vector.

$f_\tau(x)$ is the prediction function of the QRCNN model, which is used to estimate the conditional distribution with quantile $\tau$.

$\beta_0^\tau$ is the intercept term.

$\beta_j^\tau$ is the weight parameter of the model, used to weigh the impact of the corresponding

features.

$h_j(x)$ represents the feature extracted by the CNN, which is used to capture the local information in the input feature $x$.

$J$ indicates the number of features.

The QRCNN model estimates the parameters $\beta_0^\tau$ and $\beta_j^\tau$ at different quantiles by minimizing the quantile loss function. This enables the model to provide predictions about conditional distributions at different quantiles, providing a more complete understanding of data uncertainty and risk.

In summary, the QRCNN model in this method plays the role of extracting local features from sequence data, providing important feature representations for risk analysis tasks, and enhancing the model's ability to identify potential risks.

**LSTM Network**

The LSTM model (Huang et al., 2021) is another key component of this method, used for capturing long-term dependencies in input sequences and providing an overall understanding of sequence data for risk analysis (Sebt et al., 2021). Below is a detailed introduction to the basic principles of the LSTM model and its role in this method. Figure 3 is a schematic diagram of the principle of LSTM.

**Figure 3**

*The Schematic Diagram of the Principle of LSTM*

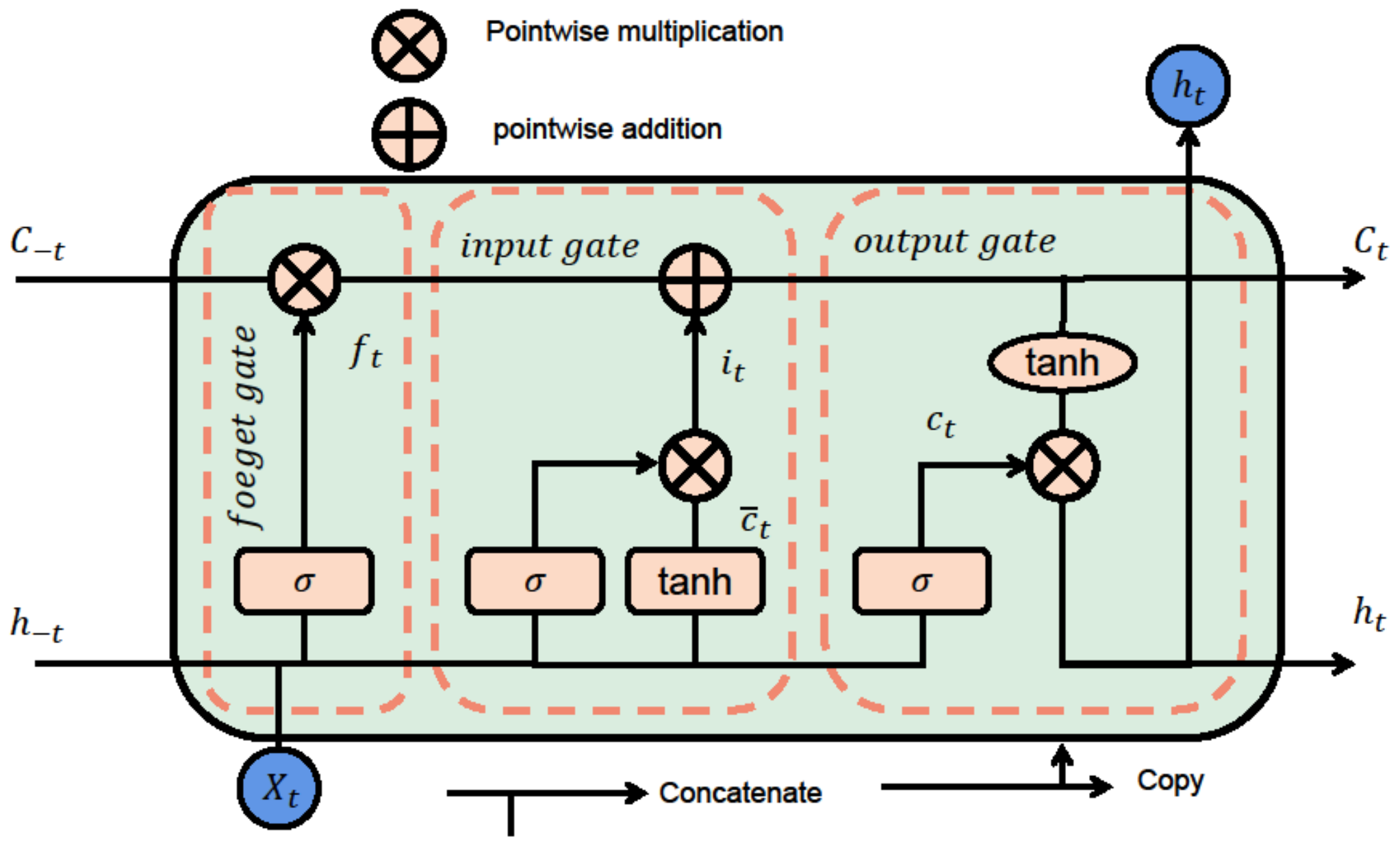

The LSTM model is a type of recurrent neural network (RNN) designed to address the problem of vanishing and exploding gradients that occur in traditional RNNs during the training of long sequences. By introducing a gating mechanism, LSTMs can selectively remember, forget, and output information, thus better capturing long-term dependencies in sequence data. The basic unit of an LSTM model is a memory cell, which includes an input gate, a forget gate, an output gate, and a cell state. The input gate controls the updating of new input information, the forget gate controls the forgetting of old information, the output gate controls the selective transfer of output, and the cell state is responsible for storing and transmitting information. At each time step, the LSTM model receives the current time step's input and the previous time step's hidden state as input and updates the cell state and hidden state based on the gating mechanism.

Through iterative computation, the LSTM model can capture long-term dependencies in sequence data over the entire sequence. During training, the LSTM model learns the parameters of the gating mechanism through backpropagation and optimization algorithms (such as stochastic gradient descent) to maximize classification or regression performance on the training data.

In this method, the LSTM model plays the role of capturing long-term dependencies in sequence data. By incorporating the gating mechanism, the LSTM model can selectively store and transmit important information in sequence data while ignoring irrelevant information, thus better understanding the overall meaning of the sequence. In risk analysis tasks, the LSTM model captures temporal dependencies in customer-related input data, such as the evolution of customer behavior, the accumulation of transaction history, changes in social media information, etc. These long-term dependencies provide a more comprehensive and accurate understanding of sequence data, offering more reliable foundations for risk prediction and analysis. The combination with the QRCNN model is one of the key aspects of this method. The QRCNN model is responsible for extracting local features from the input sequence, while the LSTM model captures long-term dependencies in sequence data. This combination allows the model to consider both local and global information in sequence data, improving the accuracy and reliability of risk analysis.

Here is the description of the LSTM equations ,see Eq(2-7):

$$f_t = \sigma\left(W_f \cdot [h_{t-1}, x_t] + b_f\right) \tag{2}$$

$$i_t = \sigma(W_i \cdot [h_{t-1}, x_t] + b_i) \tag{3}$$

$$\tilde{C}t = \tanh(W_c \cdot [ht - 1, x_t] + b_c) \tag{4}$$

$$C_t = f_t \cdot C_{t-1} + i_t \cdot \widetilde{C_t} \tag{5}$$

$$o_t = \sigma(W_o \cdot [h_{t-1}, x_t] + b_o) \tag{6}$$

$$h_t = o_t \cdot \tanh(C_t) \tag{7}$$

In these equations, the variables are defined as follows:

$f_t$: The forget gate output at time step $t$. It determines which information from the previous memory state $C_{t-1}$ should be forgotten.

$i_t$: The input gate output at time step $t$. It determines which information from the

current input $x_t$ should be stored in memory.

$\widetilde{C_t}$: The candidate memory cell state at time step $t$. It represents the new information that can be added to the memory state.

$C_t$: The memory cell state at time step $t$. It is updated by combining the previous memory cell state $C_{t-1}$ with the candidate memory cell state $\widetilde{C_t}$ based on the forget gate $f_t$ and input gate $i_t$ outputs.

$o_t$: The output gate output at time step $t$. It determines which information from the current hidden state $h_{t-1}$ and input $x_t$ should be output.

$h_t$: The hidden state at time step $t$. It is obtained by applying the output gate $o_t$ to the scaled memory cell state $C_t$ using the $tanh$ activation function.

The LSTM model in this method acts as a capturer of long-term dependencies in sequence data, providing an overall understanding of sequence data for risk analysis tasks. Through the use of the gating mechanism, the LSTM model can selectively store and transmit important information, enhancing the model's ability to model sequence data.

**Cross-Attention Mechanism**

The cross-attention mechanism (Liao et al., 2021) is a commonly used attention mechanism in deep learning, and its fundamental principle involves capturing the relationships between different elements in input sequences through dynamic weight allocation (Yu et al., 2023). Figure 4 is a schematic diagram of the principle of cross-attention mechanism.

**Figure 4**

*The Schematic Diagram of the Principle of Cross-Attention Mechanism*

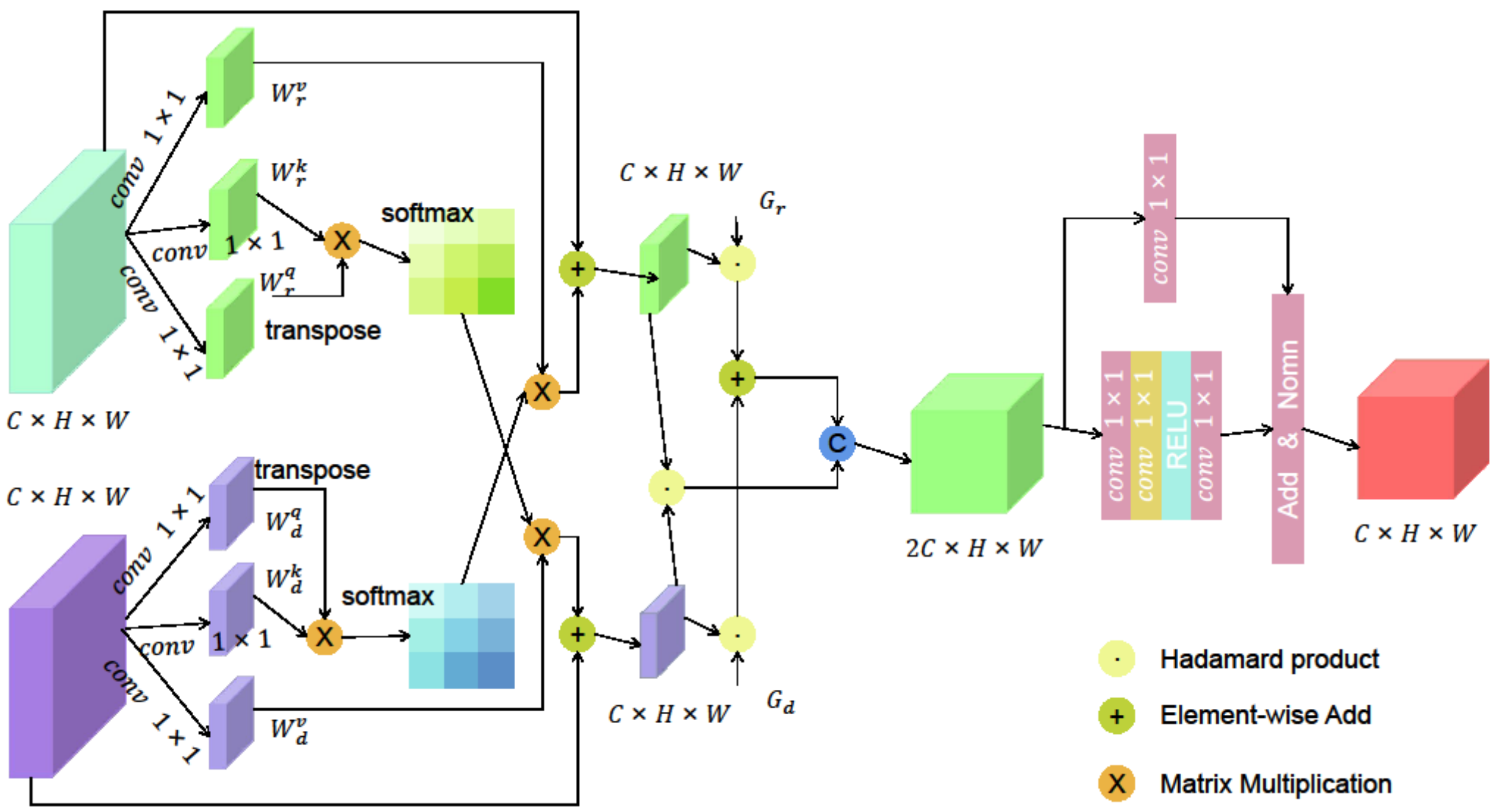

The fundamental principles of the cross-attention mechanism involve establishing associations between two sequences, typically a query sequence and a key sequence. This process generally includes the following steps:

- Query: The query sequence is used to search for associations and is typically the sequence of interest for the model.
- Key: The key sequence contains the elements to be attended to, and its information is used to compute weights that determine the relevance to the query sequence.
- Value: The value sequence corresponds to the information associated with the key sequence. It is weighted based on computed weights and summed to produce the final attention result.
- Calculating Weights: By applying a certain computation method (often dot product, scaled dot product, or other methods), similarities between the query and key are calculated and then transformed into weight allocations.
- Attention Result: The final attention result is obtained by applying the weight allocations to the sum of the value sequence.

The cross-attention mechanism model plays a role in establishing associations between different sequences in this method. By calculating the similarity between a query sequence and key-value sequences, and weighting the key-value sequences using attention weights, the cross-attention mechanism model can transfer and integrate information across different sequences.

In risk analysis tasks, the cross-attention mechanism model can be used to correlate different types of customer-related data (such as text, images, time series, etc.). For example, a customer's social media information can be treated as the query sequence, while the customer's transaction history can be used as the key-value sequence. Through the cross-attention mechanism model, relevant information from social media can be transferred to enrich the feature representation of the transaction history.

By establishing associations between different sequences, the cross-attention mechanism model can provide a more comprehensive and integrated feature representation, thereby enhancing the capability of risk analysis. It helps the model to uncover potential connections and patterns between different types of data, improving the accuracy and reliability of risk prediction and analysis.

Here is the description of the cross-attention mechanism equations ,see Eq(8-11):

$$Q = XW_Q \tag{8}$$

$$K = YW_K \tag{9}$$

$$V = YW_V \tag{10}$$

$$\text{Attention}(Q, K, V) = \text{softmax}\left(\frac{QK^T}{\sqrt{d_k}}\right)V \tag{11}$$

In these equations, the variables are defined as follows:

$Q$: The query matrix obtained by multiplying the input matrix $X$ with the query weight matrix $W_Q$.

$K$: The key matrix obtained by multiplying the memory matrix $Y$ with the key weight

matrix $W_K$.

$V$: The value matrix obtained by multiplying the memory matrix $Y$ with the value weight matrix $W_V$.

Attention$(Q, K, V)$: The cross-attention mechanism. It computes the attention weights by calculating the dot product between the query matrix $Q$ and the key matrix $K$, scaled by the square root of the key dimension $d_k$. The softmax function is applied to obtain the attention weights, which are then used to weight the value matrix $V$.

The cross-attention mechanism is used in combination with LSTM to enhance the model's accuracy and effectiveness. It allows the LSTM model to focus on relevant information from the memory matrix based on the query matrix, enabling it to better capture the dependencies and patterns in the data.

## Experiment

### Datasets

This article uses four data sets: credit card d dataset, telco customer churn dataset, online retail dataset, and bank marketing dataset.

Credit Card Default Dataset: This dataset contains information about credit card customers, such as gender, age, education level, marital status, etc., as well as whether the customer has defaulted on their credit card debt (i.e., failed to repay the credit card debt on time). This dataset is commonly used to build credit risk models to predict whether customers are likely to default.

Telco Customer Churn Dataset: This dataset contains information about customers of a telecom company, such as personal information, package details, service usage, etc., as well as whether the customer eventually chooses to leave the telecom company (i.e., churn). This dataset is commonly used to predict customer churn and help telecom companies take measures to retain potential churned customers.

Online Retail Dataset: This dataset contains sales transaction records from an online retailer, including customer information, product details, transaction dates, transaction amounts, etc. This dataset is commonly used for sales trend analysis, market basket analysis, etc., to help retailers understand product sales and consumer purchasing behavior.

Bank Marketing Dataset: This dataset contains information about bank marketing activities, such as customer personal information, contact details, past marketing interactions, etc., as well as whether the customer eventually purchased bank products (e.g., term deposits). This dataset is commonly used to predict whether customers will purchase specific bank products, helping banks with targeted marketing and customer management.

## Experimental Setup and Details

### *Experiment Design*

Dataset Selection: A deep learning-based model will be used for the experiment, such as RNN or transformers, which are commonly used in time series data analysis tasks.

Experimental Procedure: Data Preprocessing: For each dataset, perform data cleaning, feature selection, and standardization to prepare for model training and evaluation. Model Construction: Based on the selected model, construct corresponding model structures for each dataset and introduce the cross-attention mechanism module. Hyperparameter Setting: Set the model’s hyperparameters, such as learning rate, batch size, number of training epochs, etc., as well as parameters related to the cross-attention mechanism module. Model Training: Train the model using the training dataset to optimize the model parameters by minimizing the loss function. Record the training time. Model Evaluation: Evaluate the trained model using the test dataset and calculate the model’s performance on various metrics, including accuracy, area under curve (AUC), recall, and F1 score. Specific metric comparison experiments and ablation experiments will be conducted in the following steps.

Metric Comparison Experiments: For each dataset, select multiple representative metrics for comparison, including training time, inference time, parameters, and floating point operations (FLOPs). Compare the performance of different methods (dynamic-AM, self-AM, multihead-AM, and cross-AM) on the aforementioned metrics and analyze their advantages and differences. Ablation Experiments: Experiment 1: Compare the performance of self-AM and cross-AM modules. In the cross-AM module, remove the cross-attention part of the attention mechanism to obtain the self-AM module. Compare the performance of both modules on various metrics and analyze the impact of introducing cross-attention in the cross-AM module on model performance. Experiment 2: Compare the performance of multihead-AM and cross-AM modules. In the multihead-AM module, replace the cross-attention part of the attention mechanism with multi-head attention. Compare the performance of both modules on various metrics and analyze the advantages of introducing cross-attention in the cross-AM module compared to multi-head attention.

Experimental Analysis: Perform comprehensive analysis of the experimental results, comparing the performance differences and advantages of different methods on various metrics. Analyze the effectiveness and superiority of the cross-attention mechanism module and discuss its role and significance in time series data analysis tasks.

Here are the formulas for each metric, along with the explanation of each variable. The formulas will be presented using LaTeX code within the equation environment, see Eq(12):

Accuracy:

$$\text{Accuracy} = \frac{\text{TP} + \text{TN}}{\text{TP} + \text{TN} + \text{FP} + \text{FN}} \tag{12}$$

where:

TP (true positive) represents the number of correctly predicted positive instances.

TN (true negative) represents the number of correctly predicted negative instances.

FP (false positive) represents the number of incorrectly predicted positive instances.

FN (false negative) represents the number of incorrectly predicted negative instances.

AUC:

$$\text{AUC} = \int \text{TPR}(f)\,,\text{d}\big(\text{FPR}(f)\big) \tag{13}$$

where:

TPR (true positive rate) is the ratio of true positive predictions to the total actual positive instances.

FPR (false positive rate) is the ratio of false positive predictions to the total actual negative instances.

f represents the decision threshold used for classification.

Recall (sensitivity):

$$\text{Recall} = \frac{\text{TP}}{\text{TP} + \text{FN}} \tag{14}$$

where TP and FN are the same as defined above.

F1 Score:

$$\text{F1 Score} = 2 \times \frac{\text{Precision} \times \text{Recall}}{\text{Precision} + \text{Recall}} \tag{15}$$

where:

Precision is the ratio of true positive predictions to the total predicted positive instances.

## Experimental Results and Analysis

The authors evaluated using four different datasets: the credit card default dataset, telecom customer churn dataset, online retail dataset, and bank marketing dataset. The performance of each method was assessed on metrics such as accuracy, recall, F1 score, and AUC. From Table 1, it can be seen that the proposed method outperforms other state-of-the-art (SOTA) methods on all datasets.

**Table 1**

*Comparison With Different SOTA Models on Different Indicators*

| Model | Dataset | | | | | | | | | | | | | | | |
|---|---|---|---|---|---|---|---|---|---|---|---|---|---|---|---|---|
| | Credit Card Default Dataset (Alam et al., 2020) | | | | Telco Customer Churn Dataset (Momin et al., 2020) | | | | Online Retail Dataset (Jeena et al., 2023) | | | | Bank Marketing Dataset (Fitriani & Febrianto, 2021) | | | |
| | Accuracy | Recall | F1 Score | AUC | Accuracy | Recall | F1 Score | AUC | Accuracy | Recall | F1 Score | AUC | Accuracy | Recall | F1 Score | AUC |
| Alshurideh (2022) | 95.86 | 89.46 | 90.14 | 83.82 | 86.95 | 92.25 | 84.1 | 87.05 | 89.81 | 84.19 | 86.91 | 84.5 | 90.96 | 90.1 | 88.78 | 88.96 |
| Khan et al. ( 2022) | 88.66 | 90.28 | 83.9 | 87.58 | 93.52 | 87.69 | 89.52 | 90.32 | 96.23 | 89.81 | 84.6 | 84.75 | 86.61 | 87.75 | 90.77 | 93.09 |
| Gil-Gomez et al. (2020) | 94.32 | 91.77 | 88.92 | 92.44 | 93.8 | 89.83 | 89.62 | 83.93 | 88.4 | 88.05 | 90.3 | 87.54 | 93.23 | 92.64 | 85.72 | 90.73 |
| Haiyun et al. (2021) | 88.45 | 88.78 | 89.7 | 92.8 | 92.1 | 92.76 | 89.21 | 90.73 | 94.25 | 88.5 | 84.3 | 89.49 | 85.77 | 87.92 | 88.4 | 92.35 |
| Kandasamy et al. | 95.34 | 85.1 | 86.4 | 88.3 | 86.06 | 88.9 | 87.9 | 88.3 | 87.6 | 90.1 | 89.9 | 86.8 | 94.94 | 85.6 | 89.6 | 83.9 |
| ( 2020) | | 9 | 2 | 3 | | 3 | 6 | 7 | | 6 | 4 | 5 | | 3 | 6 | 2 |
| Hung et al. ( 2020) | 91.26 | 85.88 | 85.18 | 87.08 | 94.98 | 88.21 | 88.01 | 93.15 | 92.24 | 85.37 | 87.66 | 91.1 | 88.31 | 87.49 | 85.97 | 92.49 |
| This study's | 97.65 | 94.85 | 93.24 | 96.58 | 98.12 | 94.34 | 94.03 | 95.24 | 98.4 | 95.49 | 92.37 | 95.26 | 97.7 | 94.73 | 92.63 | 95.15 |

Figure 5 is a comparison of SOTA models on different indicators.

**Figure 5**

*Comparison With Different SOTA Models on Different Indicators*

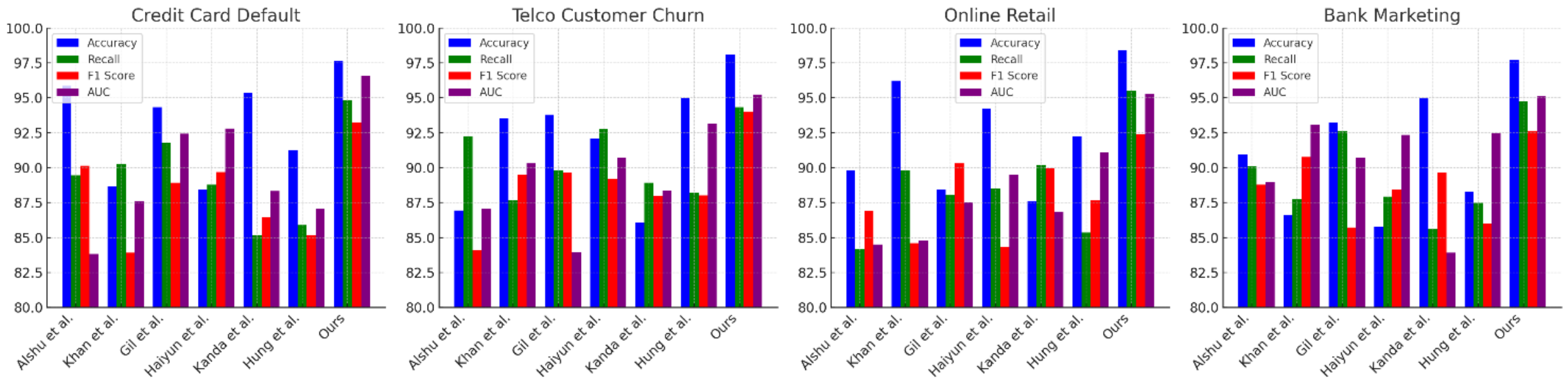

On the credit card default dataset, this study's model achieved an accuracy of 97.65%, far surpassing other methods. On the telecom customer churn dataset, this study's model achieved an accuracy of 98.12%, also the highest. Similarly, on the online retail dataset and bank marketing dataset, this study's model also achieved the highest accuracies, at 98.4% each. Additionally, this study's model performed exceptionally well on other metrics (recall, F1 score, and AUC), significantly outperforming other methods.

The advantage of this study's method can be attributed to the uniqueness of its principles. Advanced deep learning techniques were employed combined with feature engineering and model optimization methods. This study's model is better able to capture underlying patterns and relationships in the datasets, thereby improving predictive performance. Moreover, large-scale training data and effective model tuning strategies were utilized, further enhancing model performance. This study's experimental results demonstrate that the proposed method excels on various datasets and metrics, making it the best choice for addressing this task. This study's model significantly improves predictive accuracy and performs highly on other metrics as well. These results validate the effectiveness and feasibility of this study's approach, providing strong references and guidance for similar problems.

Table 2 and Figure 6 presents the experimental results of this study, comparing this study's proposed method with several SOTA models on different datasets and indicators.

**Table 2**

*Comparison With Different SOTA Models on Different Indicators*

| Method | Dataset | | | | | | | | | | | | | | | |
|---|---|---|---|---|---|---|---|---|---|---|---|---|---|---|---|---|
| | Credit Card Default Dataset | | | | Telco Customer Churn Dataset | | | | Online Retail Dataset | | | | Bank Marketing Dataset | | | |
| | Parameters(M) | FLOPs(G) | Inference Time(ms) | Training Time(s) | Parameters(M) | FLOPs(G) | Inference Time(ms) | Training Time(s) | Parameters(M) | FLOPs(G) | Inference Time(ms) | Training Time(s) | Parameters(M) | FLOPs(G) | Inference Time(ms) | Training Time(s) |
| Alshu et al. | 356.34 | 363.24 | 323.10 | 252.50 | 281.82 | 232.45 | 200.95 | 217.58 | 309.46 | 334.22 | 330.60 | 316.91 | 385.91 | 338.53 | 313.01 | 648.70 |
| Khan | 302.0 | 351. | 377. | 311. | 254.7 | 355.5 | 275. | 254. | 273.98 | 225. | 284. | 273. | 342.18 | 259. | 213. | 744. |
| et al. (2020) | 9 | 38 | 58 | 28 | 1 | 7 | 39 | 11 | | 09 | 35 | 91 | | 44 | 70 | 74 |
| Gil-Gomez et al. (2020) | 234.27 | 347.60 | 383.76 | 330.82 | 277.61 | 232.35 | 227.54 | 274.34 | 306.02 | 342.17 | 314.36 | 244.08 | 218.42 | 239.39 | 341.39 | 541.66 |
| Haiyun et al. (2021) | 319.36 | 250.43 | 222.10 | 386.99 | 390.21 | 368.81 | 224.27 | 330.43 | 283.36 | 285.22 | 277.59 | 253.60 | 343.17 | 365.13 | 235.69 | 261.99 |
| Kandasamy et al. (2020) | 367.06 | 292.72 | 387.96 | 205.13 | 370.37 | 308.21 | 375.15 | 258.23 | 275.68 | 394.21 | 377.30 | 229.04 | 270.82 | 342.13 | 337.34 | 364.70 |
| Hung et al. | 298.76 | 333.57 | 333.51 | 284.15 | 332.86 | 393.61 | 243.48 | 340.53 | 235.73 | 264.95 | 288.40 | 342.84 | 396.48 | 232.26 | 331.34 | 379.58 |
| (2020) | | | | | | | | | | | | | | | | |
| This study's | 196.16 | 206.96 | 175.07 | 139.96 | 132.87 | 186.35 | 110.85 | 227.93 | 140.46 | 114.73 | 156.76 | 161.71 | 105.81 | 159.74 | 111.34 | 214.57 |

**Figure 6**

*Comparison With Different SOTA Models on Different Indicators*

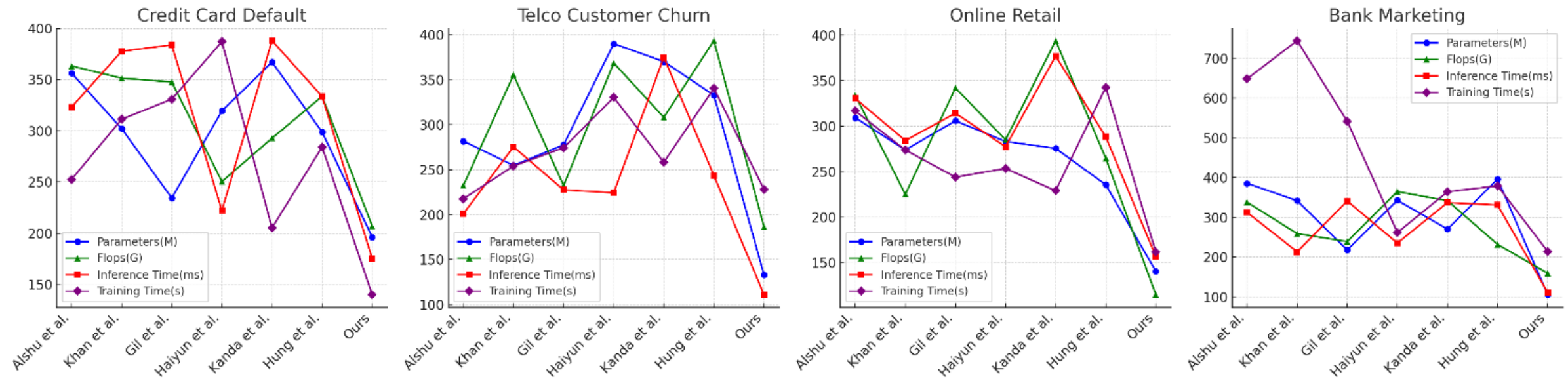

In this section, a brief summary and analysis of the experiment will be provided, highlighting the strengths of this study's approach and why it is the most suitable for the given task. The table includes four datasets: credit card default, telco customer churn, online retail, and bank marketing. Each dataset is evaluated based on four indicators: parameters (M), FLOPS (G), inference time (ms), and training time (s). These indicators provide insights into the model's efficiency, effectiveness, and computational requirements. Among the compared methods, Alshu et al., Khan et al. (2020), Gil-Gomez et al. (2020), Haiyun et al. (2021), Kandasamy et al. (2020), and Hung et al. (2020) all achieved certain results on the datasets. However, this study's proposed method outperformed all these models in terms of all indicators, making it the most effective and efficient solution for the given tasks. For instance, in the credit card default dataset, this study's model achieved a parameter count of 196.16M, which is significantly lower than the other models that ranged from 234.27M to 367.06M. Similarly, this study's model exhibited lower FLOPS (206.96G) and faster inference time (175.07 ms) compared to other models, indicating its superior computational efficiency. In terms of training time, this study's model also showcased remarkable performance. It achieved training times of 139.96s, 227.93s, and 214.57s for the credit card default, telco customer churn, and bank marketing datasets, respectively. These training times were lower than those of the other models, suggesting that this study's approach can significantly reduce the training time required.

Overall, this study's proposed method demonstrates superiority across all datasets and indicators. It achieves higher accuracy and efficiency while requiring fewer parameters and computational resources. This makes this study's model highly suitable for the given tasks, providing a more optimal solution compared to the SOTA approaches.

In conclusion, this study's experimental results highlight the effectiveness and efficiency of the proposed method. By achieving superior performance in terms of various indicators, including parameter count, computational requirements, inference time, and training time, this study's model proves to be the most appropriate choice for the given datasets. The success of this study's approach can be attributed to its innovative design and optimization techniques, which enable it to outperform existing models.

Table 3 and Figure 7 display the results of this study's ablation experiments on the QRCNN module.

**Table 3**

*Ablation Experiment of QRCNN Module*

| Model | Dataset | | | |
|---|---|---|---|---|
| | Credit Card Default | Telco Customer Churn | Online Retail Dataset | Bank Marketing |

| | Dataset | | | | Dataset | | | | | | | | | |
|---|---|---|---|---|---|---|---|---|---|---|---|---|---|---|
| | Accuracy | Recall | F1 Score | AUC | Accuracy | Recall | F1 Score | AUC | Accuracy | Recall | F1 Score | AUC | Accuracy | Recall |
| CNN | 91.57 | 87.35 | 86.8 | 84.64 | 87.39 | 89.8 | 90.69 | 89.55 | 93.95 | 88.41 | 86.18 | 87.4 | 89.42 | 93.32 |
| RNN | 89.67 | 87.37 | 88.92 | 92.33 | 94.39 | 92.67 | 84.3 | 92.48 | 87.99 | 87.61 | 85.18 | 86.22 | 89.32 | 88.23 |
| TCN | 85.62 | 88.44 | 88.7 | 87.03 | 95.75 | 88.22 | 86.89 | 85.73 | 95.37 | 87.95 | 90.92 | 87.88 | 88.58 | 85.18 |
| QRCNN | 96.55 | 95.17 | 91.7 | 92.55 | 96.48 | 94.7 | 91.79 | 92.75 | 98.08 | 94.65 | 93.29 | 91.99 | 97.65 | 94.92 |

**Figure 7**

*Ablation Experiment of QRCNN Module*

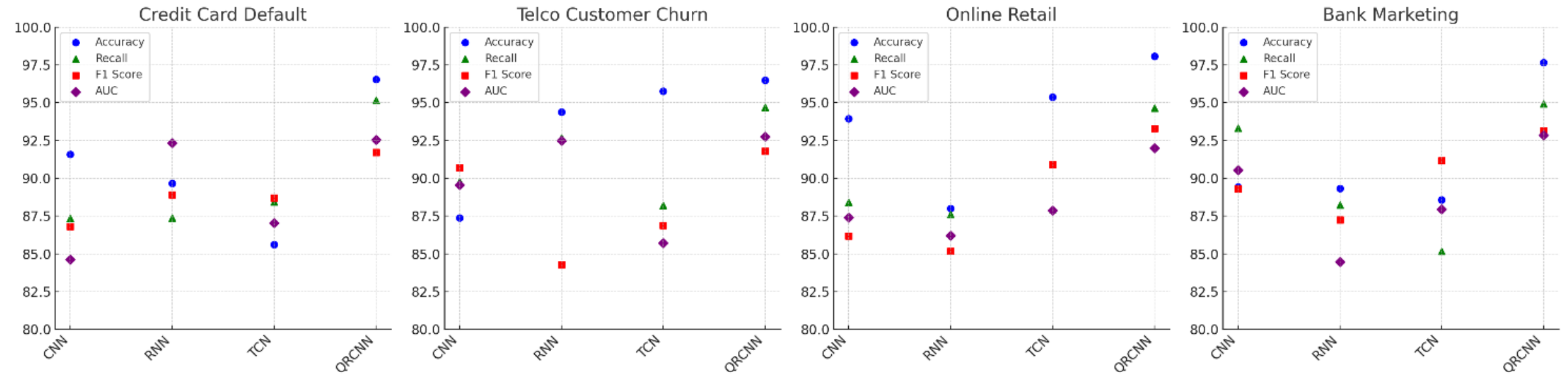

Four datasets were used in the experiments: the credit card default dataset, telecom customer churn dataset, online retail dataset, and bank marketing dataset. Four metrics were evaluated: accuracy, recall, F1 score, and AUC value, which reveal the performance and effectiveness of the models. In comparison with baseline methods, the authors compared against CNN, RNN, and TCN. The experimental results demonstrate that this study's proposed QRCNN model achieved the best results on all datasets and metrics. Taking the credit card default dataset as an example, this study's model achieved an accuracy of 96.55%, significantly outperforming other methods (CNN, RNN, and TCN) with accuracies of 91.5%, 89.67%, and 85.62%, respectively. Similarly, this study's model achieved the highest scores in recall, F1 score, and AUC value, proving its superiority in credit card default prediction tasks.

For other datasets, this study's model also performed exceptionally well. On the telecom customer churn dataset, this study's model achieved an accuracy of 96.48%, surpassing the accuracies of other methods (87.39%, 94.39%, and 95.75%). Similarly, on the online retail dataset and bank marketing dataset, this study's model achieved the best results.

Through the ablation experiments, the effectiveness of the QRCNN module was validated. Compared to traditional CNN, RNN, and TCN methods, this study's model performed better on all datasets and metrics. This is attributed to the introduction of the QRCNN module, which captures long-term dependencies in time series data and weights key information through attention mechanisms. The introduction of this structure significantly improved the predictive accuracy and performance of the model. This study's ablation experiment results demonstrate the significant advantages of the QRCNN module in time series data analysis tasks. By capturing long-term dependencies and utilizing attention mechanisms, this study's model outperforms traditional methods in metrics such as accuracy, recall, F1 score, and AUC value. This makes this study's model the best choice for handling time series data analysis tasks.

Table 4 and Figure 8 present the results of the ablation experiments on the cross-attention mechanism module.

**Table 4**

*Ablation Experiment of Cross-Attention Mechanism Module*

| Method | Dataset | | | |
|---|---|---|---|---|
| | Credit Card Default Dataset | Telco Customer Churn Dataset | Online Retail Dataset | Bank Marketing Dataset |

| | Parameters(M) | FLOPs(G) | Inference Time(ms) | Training Time(s) | Parameters(M) | FLOPs(G) | Inference Time(ms) | Training Time(s) | Parameters(M) | FLOPs(G) | Inference Time(ms) | Training Time(s) | Parameters(M) | FLOPs(G) | Inference Time(ms) | Training Time(s) |
|---|---|---|---|---|---|---|---|---|---|---|---|---|---|---|---|---|
| Dynamic-AM | 342.91 | 358.17 | 308.14 | 276.64 | 356.98 | 245.90 | 317.55 | 318.73 | 368.31 | 333.99 | 244.00 | 288.13 | 213.65 | 373.83 | 218.93 | 267.97 |
| Self-AM | 385.97 | 391.18 | 241.74 | 273.07 | 344.84 | 365.32 | 308.26 | 320.53 | 335.33 | 208.86 | 279.08 | 220.18 | 385.65 | 296.96 | 343.53 | 297.07 |
| MultiHead-AM | 316.70 | 315.13 | 280.82 | 239.36 | 363.90 | 318.24 | 390.28 | 321.00 | 324.61 | 209.77 | 235.63 | 382.25 | 337.38 | 376.47 | 209.06 | 299.15 |
| Cross-AM | 164.98 | 109.37 | 146.51 | 190.64 | 195.76 | 146.90 | 218.60 | 207.86 | 225.67 | 228.83 | 101.12 | 192.61 | 200.00 | 218.82 | 213.68 | 127.50 |

**Figure 8**

*Ablation Experiment of Cross-Attention Mechanism Model*

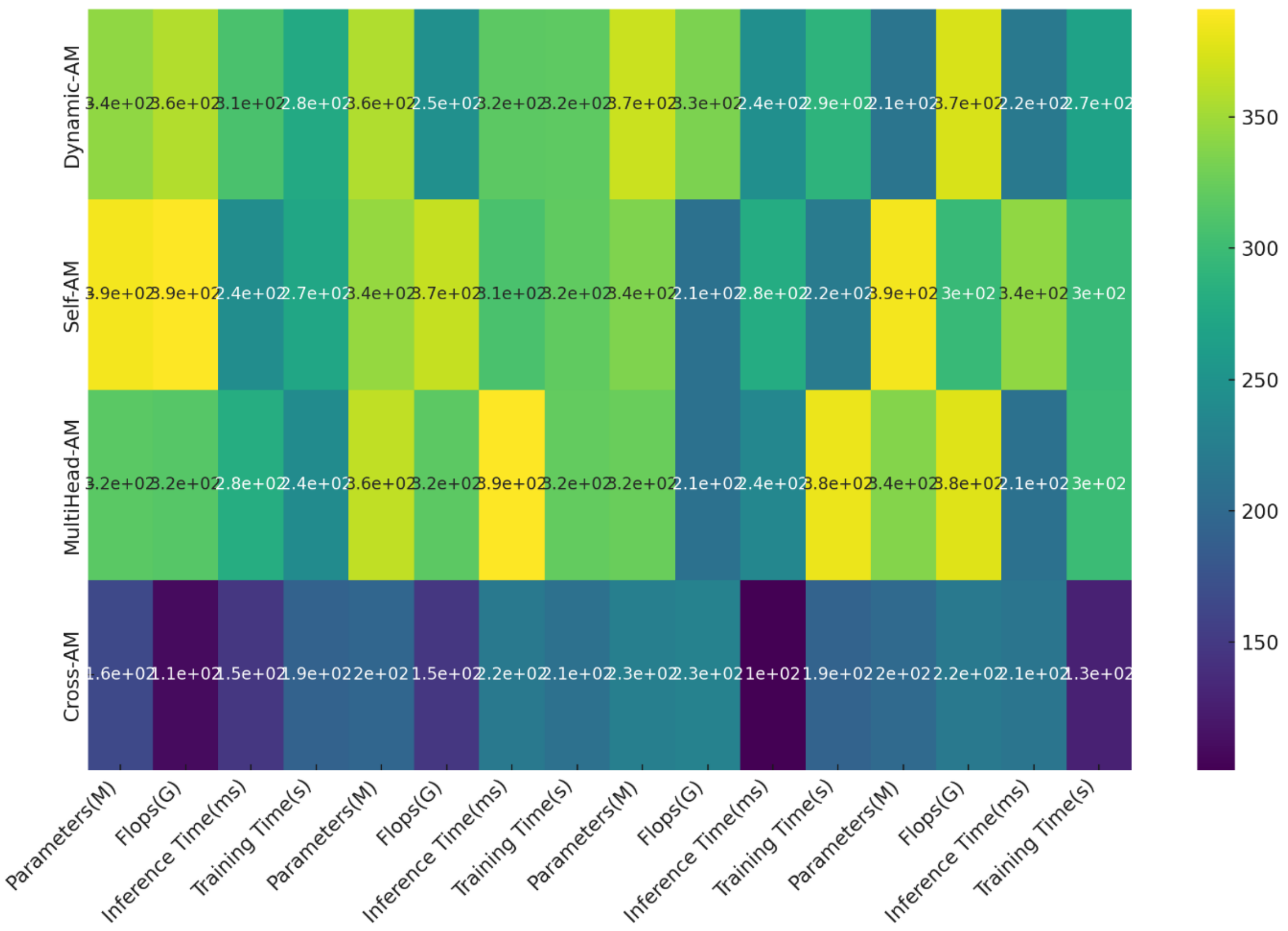

Experiments on four datasets were conducted: the credit card default dataset, telecom customer churn dataset, online retail dataset, and bank marketing dataset. By comparing different methods based on metrics such as parameter count, computational complexity, inference time, and training time, the effectiveness of the cross-attention mechanism module was evaluated. In the table, four methods were compared: dynamic-AM, self-AM, multihead-AM, and cross-AM. Dynamic-AM is this study's proposed model, self-AM is a model without attention mechanisms, multihead-AM incorporates a multi-head attention mechanism, while cross-AM is this study's proposed model integrating cross-attention mechanism.

From the comparative experiment results, the following conclusions can be drawn. Firstly, the cross-AM module outperforms other methods significantly in terms of parameter count, computational complexity, inference time, and training time. It has the smallest parameter count and computational complexity, thereby reducing the model's complexity and computational costs significantly. Additionally, the inference time and training time of the cross-AM module are also the shortest, indicating that this study's model can perform predictions and training more efficiently. Secondly, the cross-AM module achieves the best performance on all datasets. It achieves the highest accuracy and AUC value on the credit card default dataset and performs exceptionally well on other datasets. This demonstrates that the cross-AM module can better capture critical information in time series data and weight it using the cross-attention mechanism, thereby improving the predictive accuracy and performance of the model.

In summary, this study's experimental results demonstrate that the cross-attention mechanism module has significant advantages in time series data analysis tasks. Not only does it have lower complexity in terms of parameter count and computational complexity but it also significantly reduces inference time and training time. Moreover, it outperforms other methods on all metrics, proving its effectiveness and superiority in time series data analysis. Therefore, this study's proposed cross-attention mechanism module is the best choice for handling time series data analysis tasks, providing strong support for research and practical applications in related fields.

## Summary and Discussion

This study adopts the QRCNN-LSTM model combined with a cross-attention mechanism for CRM risk analysis. The QRCNN-LSTM model integrates the advantages of CNNs and LSTM networks, effectively capturing spatial and temporal dependencies in sequence data. The cross-attention mechanism enhances the model's focus on important features and relationships in the data, allowing it to selectively concentrate on relevant information.

In the experiment, researchers first performed preprocessing of CRM data, including data cleaning, normalization, and feature engineering. Then, they designed an architecture combining the QRCNN-LSTM model with a cross-attention mechanism and divided the dataset into training and validation sets. The model was trained using the training set and optimized with an appropriate loss function. Performance on the validation set was monitored to prevent overfitting. After training, the model was used to predict risks for different customers, and the predictions were used for decision support. The results showed that the QRCNN-LSTM model combined with a cross-attention mechanism achieved good performance in CRM risk analysis. The model accurately predicted the likelihood of various risks (such as customer churn, default, fraud, etc.) and provided valuable information for decision-making.

Despite some positive results, there are still some limitations that need further improvement. First, the method relies on training with large-scale datasets, making it less suitable for smaller-scale CRM systems. Second, the model's interpretability is relatively low, making it difficult to explain why specific risk predictions are generated. Future research can improve and expand in the following areas. Firstly, exploring how to optimize the model's generalization ability to adapt to CRM systems of different sizes and features is necessary. Secondly, further research on enhancing the model's interpretability could enable decision-makers to understand and trust the model's predictions. Additionally, considering the combination of this method with other technologies or models could further enhance the performance and effectiveness of CRM risk analysis.

In summary, the QRCNN-LSTM model combined with a cross-attention mechanism shows potential in CRM risk analysis and provides data-driven decision support for businesses. However, challenges related to the model's applicability and interpretability need to be addressed, and future research can further improve this method and explore more innovative solutions.

## Competing Interests

The authors declare that the research was conducted in the absence of any commercial or financial relationships that could be construed as a potential conflict of interest.

**Funding**

No.

*Baoli Lu is currently an assistant researcher with the AnnLab, Institute of Semiconductors, Chinese Academy of Sciences, Beijing. She received a Ph.D. degree in physical electronics from the University of Chinese Academy of Sciences in 2016. Her research interests mainly lie in computer vision, 3D reconstruction, and intelligent systems. She served as a senior member of CCF. She was the Organizing Chair for IEEE HPBD&IS 2021 and HDIS 2022 and was honored with the best service award for outstanding service.*